\pdfoutput=1

\documentclass[11pt]{article}

\usepackage{acl}

\usepackage{times}
\usepackage{latexsym}

\usepackage[T1]{fontenc}

\usepackage[utf8]{inputenc}

\usepackage{microtype}

\usepackage{todonotes}
\usepackage{amsmath}
\usepackage{siunitx}
\usepackage{tikz,pgfplots}
\pgfplotsset{compat=1.16}

\title{AppTek's Submission to the IWSLT 2022 \\ Isometric  Spoken Language Translation Task}

\author{Patrick Wilken \\
  AppTek  \\
  Aachen, Germany \\
  \texttt{pwilken@apptek.com} \\\And
  Evgeny Matusov \\
  AppTek  \\
  Aachen, Germany \\
  \texttt{ematusov@apptek.com} \\}

\begin{document}
\maketitle
\begin{abstract}
To participate in the Isometric Spoken Language Translation Task of the IWSLT 2022 evaluation, 
constrained condition, AppTek developed neural Transformer-based systems for English-to-German with various mechanisms of length control, ranging from source-side and target-side pseudo-tokens to encoding of remaining length in characters that replaces positional encoding.
We further increased translation length compliance by sentence-level selection of length-compliant hypotheses from different system variants, as well as rescoring of N-best candidates from a single system. Length-compliant back-translated and forward-translated synthetic data, as well as other parallel data variants derived from the original MuST-C training corpus were important for a good quality/desired length trade-off. Our experimental results show that length compliance levels above 90\% can be reached while minimizing losses in MT quality as measured in BERT and BLEU scores.
\end{abstract}

\section{Introduction}
In this paper, we describe AppTek's submission to the IWSLT 2022 Isometric Spoken Language Translation evaluation \citep{iwslt:2022}. Our goal was to create a system that produces translations which are within 10\% of the source sentence length, but have similar levels of quality as a baseline system translations without length control. AppTek participated in the constrained condition with an English-to-German neural machine translation (NMT) system that we describe in Section~\ref{sec:base}. The system was extended with 5 different length control methods, which we explain in detail in Section~\ref{sec:length_control}. We also created synthetic data with back-translation, forward-translation, as well as a novel data augmentation method of synonym replacement. All three methods are described in Section~\ref{sec:synthetic}. Our experimental results on the MuST-C tst-COMMON test set and the official evaluation test set are presented in Section~\ref{sec:experiments}, including ablation studies that prove the effectiveness of synthetic data and noisy length encoding for a better trade-off between length compliance and MT quality. We summarize our findings in Section~\ref{sec:conclusions}.

\section{Baseline system}\label{sec:base}

\subsection{Data}
We follow the constrained condition of the IWSLT Isometric SLT task and use only English-to-German TED-talk data from the MuST-C corpus \citep{di2019must}. The corpus contains 251K sentence pairs with 4.7M and 4.3M English and German words, respectively.

We apply minimal text pre-processing, mainly consisting of normalization of quotes and dashes. 2K sentences that have mismatching digits or parentheses in source and target were filtered out.

We use a joint English and German SentencePiece model \cite{kudo-richardson-2018-sentencepiece}, trained on the whole corpus using a vocabulary size of 20K, to split the data into subwords.

\subsection{Neural NMT model}
In preliminary experiments we tried several Transformer model configurations, including \textit{base} and \textit{big} from the original paper \citep{vaswani2017attention}, a 12 encoder and decoder layer variant of \textit{base}, and a "deep" 20 encoder layer version with halved feed-forward layer dimension in the encoder and only 4 attention heads. These attempts to optimize the model architecture for the given, rather low resource task did not yield a better architecture than Transformer \textit{big}, which we end up using in all our experiments.

We however find an increased dropout rate of 0.3 and an increased label smoothing of 0.2 to be crucial.
We further optimize the model by sharing the parameters of the source and target embeddings as well as the softmax projection matrix.

In all experiments we use two translation factors \citep{garcia-martinez-etal-2016-factored} on both the source and target side to represent the casing of the subwords and the binary decision whether a subword is attached to the previous subword \cite{wilken2019novel}. This allows for explicit sharing of information between closely related variants of a subword and reduces the model vocabulary size.

All models are trained on a single GPU for 162 to 198 epochs of 100K sentence pairs each in less than two days. We use batches of 1700 subwords and accumulate gradients over 8 subsequent batches. The global learning rate of the Adam optimizer is increased linearly from \num{3e-5} to \num{3e-4} in the first 10 epochs and then decreased dynamically by factor 0.9 each time perplexity on the MuST-C dev set increases during 4 epochs. For decoding we use beam search with a beam size of 12.

We train the Transformer models using \mbox{RETURNN} \citep{doetsch2017returnn, zeyer-etal-2018-returnn}, which is a flexible neural network toolkit based on Tensorflow \citep{tensorflow2015-whitepaper}. Automation of the data processing, training and evaluation pipelines is implemented with Sisyphus \citep{peter-etal-2018-sisyphus}.

\section{Length control methods}
In this work we perform an extensive evaluation of different ways to control the length of the translations generated by the NMT model, all applied to the same baseline Transformer \textit{big} model.

\label{sec:length_control}
\subsection{N-best rescoring}
A simple method to achieve length compliant translation is to generate N-best lists and select translation hypotheses from the lists that adhere to the desired length constraints. \citet{saboo-baumann-2019-integration} and \citet{lakew2021machine} compute a linear combination of the original MT model score and a length-related score to reorder the N-best list. In this work, we simply extract the translation from the N-best list with the best MT score that has a character count within a 10\% margin of the source character count and fall back to the first best hypothesis if there is no such translation. This approach is tailored towards the evaluation condition of the IWSLT Isometric SLT task where length compliance within a 10\% margin is a binary decision and the absolute length difference is not considered.

While N-best rescoring has the advantage of being applicable to any NMT model that uses beam search, it is outperformed by learned length control methods because in many cases there is no length compliant translation in the N-best list, and also because learned methods are able to shorten the translation in a more semantically meaningful way. However, we use N-best rescoring on top of other methods to further improve length compliance, as done by \citet{lakew2021machine}.

\subsection{Length class token}
\citet{lakew-etal-2019-controlling} introduce a special token at the start of the source sentence to control translation length. For this, the training data is classified into difference length classes based on the target-to-source ratio measured in number of characters.
In this work we use two variants of length classes:
\begin{enumerate}
    \item 3 length bins representing "too short", "length compliant" and "too long". Length compliant here means the number of characters in source and target differs by less than 10\%;
    \item 7 length bins from "extra short" to "extra long", such that an approximately equal number of training sentence pairs falls into each bin.
\end{enumerate}
 
The first option is focused on isometric MT, i.e. equal source and target length, while the second option offers a more fine-grained length control.

In addition, we analyze the difference of adding the token to the source versus the target side. Adding the token on the target side has the advantage of offering the option to not enforce a length class at inference time and instead let the model perform an unbiased translation. This is especially important in a commercial setting where costs can be saved by deploying a single model for general and isometric MT. 

\subsubsection{Length ROVER}
\label{subsubsec:rover}
A system that takes a length class as input can produce multiple different translations of a given source sentence. To maximize the chance for length compliant translations, we produce translations of the whole test set for each of the length bins and then, for each sentence, select the hypothesis which adheres to the length constraint. We refer to this as length ROVER, in analogy to the automatic speech recognition system combination technique called ROVER~\cite{ROVER}. If multiple length bins produce a length compliant translation, precedence is determined by the corpus-level translation quality scores for the different length bins. If no bin produces a length compliant translation the bin with the best corpus-level translation quality is used as fallback. 

As we use a target-side length token, we can let the model predict the length token instead of forcing one. This usually leads to the best corpus-level translation quality. We include this freely decoded translation in the length ROVER.

When applying the length ROVER to the 7-bin model, we exclude the bins corresponding to the longest and shortest translations as those rarely lead to length compliant translations but generally to degraded translation quality. The same is true for the "too short" and "too long" bins in the 3-bin model, which is why we do not use the length ROVER for this model.

\subsection{Length encoding}
\label{subsec:length_encoding}
We adopt length-difference positional encoding (LDPE) from \citet{takase-okazaki-2019-positional}. It replaces the positional encoding in the transformer decoder, which usually encodes the absolute target position, with a version that "counts down" from a desired output length $L_\mathrm{forced}$ to zero. At each decoding step the available remaining length is an input to the decoder and thus the model learns to stop at the right position. In training, $L_\mathrm{forced}$ is usually set to the reference target length $L_\mathrm{target}$, while at inference time it can be set as desired. For isometric MT, setting it to the source length $L_\mathrm{forced} = L_\mathrm{source}$ is the natural choice.

The original work of \citet{takase-okazaki-2019-positional} uses a character-level decoder, which means that the number of decoding steps equals the translation length, assuming the latter is measured in number of characters. Using subwords \cite{sennrich-etal-2016-neural} as the output unit of the decoder is more common in state-of-the-art systems \citep{akhbardeh-etal-2021-findings}. In this case, one can either encode the target length in terms of number of subword tokens \citep{liu-etal-2020-adapting, niehues-2020-machine, buet2021toward}, or keep the character-level encoding which however requires subtracting the number of characters in the predicted subword token in each decoding step \citep{lakew-etal-2019-controlling}. The former has the disadvantage that the number of subword tokens is a less direct measure of translation length, especially for the case of the IWSLT Isometric SLT task where length compliance is measured in terms of number of characters. The second option is more exact but arguably a bit more complex to implement. In this work we compare results for both methods.

In contrast to \citep{lakew-etal-2019-controlling} we do not combine standard token-level positional encoding and character-level length encoding, instead we only use the latter.

\subsubsection{Length perturbation}
\label{subsubsec:length_perturbation}

For both the token-level and character-level version we add random noise to the encoded translation length $L_\mathrm{forced}$ during training \citep{oka-etal-2020-incorporating}. We find that this is necessary to make the model robust to the mismatch between training, where the target length is taken from a natural translation, and inference, where the enforced target length is a free parameter. Especially in the case of character-length encoding one cannot expect that a high-quality translation with a given exact character count exists.
As opposed to \citet{oka-etal-2020-incorporating}, who add a random integer to the token-level target length sampled from a fixed interval, e.g. $[-4, 4]$, we chose a relative +/-10\% interval:
\begin{equation}
    L_\mathrm{forced} \sim U\left(\lfloor 0.9 \cdot L_\mathrm{target} \rceil, \lfloor 1.1 \cdot L_\mathrm{target} \rceil \right)
\end{equation}

Here, $U(n,m)$ denotes the discrete uniform distribution in the interval $[n,m]$, and $\lfloor \cdot \rceil$ denotes rounding to the nearest integer. This is in line with the +/-10\% length compliance condition used in the evaluation. The length difference subtracted in each decoder step is left unaltered, which means counting down will stop at a value that in general is different from zero. 

\subsubsection{Second-pass length correction}

Length encoding as described above does not result in a length compliant translation in all cases. The reasons for this are: 1. general model imperfections, intensified by the small size of the training data in the constrained track; 2. the noise added to the target length in training (although it is within the "allowed" 10\% range); 3. for the case of token-level length encoding, an equal number of source and target tokens does not necessarily mean an equal number of characters.

We therefore perform a second decoding pass for those sentences where the first pass does not generate a length compliant translation. In this second pass, instead of attempting to enforce $L_\mathrm{forced} = L_\mathrm{source}$, we make a correction by multiplying by the source-to-target ratio observed in the first pass (measured in tokens or characters, depending on the unit used for length encoding):
\begin{equation}
    L^\mathrm{2\text{-}pass}_\mathrm{forced} = \left\lfloor L_\mathrm{source} \cdot \frac{L_\mathrm{source}}{L^\mathrm{1\text{-}pass}_\mathrm{target}}\right\rceil
\end{equation}
$L^\mathrm{1\text{-}pass}_\mathrm{target}$ is the first pass translation length, $\lfloor \cdot \rceil$ denotes rounding.
That way, an over-translation of factor $r$ in the first pass will be counteracted by "aiming" at a translation length of $1 / r$ of the source length in the second pass.

This procedure could be applied iteratively, one could even run a grid search of many different values for $L_\mathrm{forced}$ until a length compliant translation is generated. We refrain from doing so as we find it to be impracticable in real-world applications.

\section{Synthetic data}
\label{sec:synthetic}
We expand the original MuST-C data with synthetic data of different types, all derived from the given MuST-C corpus.

First, we include a copy of the data\footnote{Including, if applicable, the synthetic data described below.} in which two consecutive sentences from the same TED talk are concatenated into one.
Since many segments in the original data are short, this helps to learn more in-context translations. Then, we also include a copy of the data where the English side is preprocessed by lowercasing, removing punctuation marks and replacing digits, monetary amounts and other entities with their spoken forms. This helps to adjust to the spoken style of TED talks and imperfections in the (manual) transcriptions of the training and evaluation data. 

We also use 82K bilingual phrase pairs extracted from word-aligned MuST-C data, as described below, as training instances. 

\subsection{Word synonym replacement}
\label{subsec:synonyms}
To enrich the training data with more examples of length-compliant translations, we experiment with a novel technique of replacing a few randomly selected source (English) words in a given sentence pair with their synonyms which are shorter/longer in the number of characters, so that the resulting modified synthetic sentence is closer to being length compliant. Whereas in an unconstrained conditions the synonyms can come from WordNet or other sources, in the constrained track we rely on synonyms extracted from a bilingual lexicon. The replacement of a source word with a synonym in a given sentence pair happens only if it is aligned to a target word, for which another word translation exists in the bilingual lexicon.

The word alignment and bilingual word lexicon extraction is performed on the lowercased MuST-C corpus itself using FastAlign~\cite{dyer-etal-2013-simple}. The bilingual lexicon is filtered to contain entries with the costs (negative log of the word-level translation probability) of 50 or lower.

We apply the synonym replacements only to sentence pairs for which the target sentence is not length-compliant with the source. We first generate multiple versions of modified source sentences for these data, which all differ in the choice of randomly selected words that are to be replaced with synonyms and in the actual synonyms selected for replacement (also at random). Each word in a sentence has a 0.5 chance of being considered for replacement (regardless of whether it has synonyms as defined above or not), and the replacement is done with (at most) one of 3 synonym candidates with the highest lexicon probability which have fewer or more characters than the word being replaced, depending on whether the length of the original sentence was too long or too short.

From the resulting data (ca.~1M sentences), we keep only those modified source sentences for which the BERT F1 score \citep{bert-score} with respect to the original (unmodified) source sentence is 0.94 or higher. In this way we try to make sure that the meaning of the modified source sentence stays very close to the original meaning. This way, only 192K sentences are kept, which are then paired with the original target (German) sentences to form a synthetic synonym replacement parallel corpus.

\subsection{Back-translated data}
We train the reverse, German-to-English system with 7 length bins and source length token as described in Section~\ref{sec:length_control} using the same architecture and settings as for the English-to-German system. We then use this system to translate the MuST-C corpus from German to English, generating 7 translations of each sentence for each of the 7 bins. From these data, we keep all back-translations which make the corresponding German sentence length-compliant. This resulted in a back-translated corpus of 172K sentence pairs.

\begin{table*}[ht]
\small
\centering
\begin{tabular}{| c | l | c c c | c c c |}
 \hline
 & & \multicolumn{3}{|c|}{\textbf{tst-COMMON v2}} & \multicolumn{3}{|c|}{\textbf{blind test}} \\
 \# & & \textbf{BLEU} & \textbf{BERT} & \textbf{LC} & \textbf{BLEU} & \textbf{BERT} & \textbf{LC} \\
 \hline
 0 & \textbf{baseline} (\textit{no length control}) & 32.0 & 84.00 & 44.03 & 19.2 & 77.94 & 45.50 \\
 \hline
 1 & \textbf{source-side token, 3 bins} & 31.3 & 83.94 & 51.59 & 20.6 & 78.40 & 62.50 \\
 2 & \quad + N-best rescoring & 30.5 & 83.60 & 78.41 & 20.1 & 77.78 & 81.50 \\
 \hline
 3 & \textbf{target-side token, 3 bins}  & 31.4 & 83.88 & 50.12 & 19.7 & 78.37 & 53.50  \\
 4 & \quad + N-best rescoring & 30.7 & 83.58 & 77.40 & 18.3 & 77.43 & 82.50 \\
 \hline
 & \textbf{target-side token, 7 bins} & & & & & & \\
 5 & predicted token (\textit{no length control}) & 32.0 & 84.00 & 45.23 & 18.3 & 77.55 & 46.50 \\
 6 & \quad + N-best rescoring  & 31.1 & 83.75 & 71.20 & 18.9 & 77.38 & 72.50 \\
 7 & M token  & 31.7 & 83.99 & 49.19 & 19.1 & 78.24 & 56.00 \\
 8 & \quad + N-best rescoring  & 31.0 & 83.74 & 76.39 & 18.6 & 77.68 & 81.00 \\
 9 & S token  & 30.5 & 83.73 & 62.95 & 18.9 & 78.05 & 59.00 \\
 10 & \quad + N-best rescoring  & 29.8 & 83.38 & 87.64 & 18.9 & 77.52 & 85.50 \\
 11 & XS token  & 28.1 & 83.09 & 72.13 & 18.2 & 77.81 & 68.00 \\
 12 & \quad + N-best rescoring  & 27.8 & 82.91 & 92.21 & 17.8 & 77.32 & 90.00 \\
 13 & ROVER over XS to XL  & 29.0 & 83.35 & 80.66 & 17.5 & 77.59 & 76.50 \\
 14 & \quad + N-best rescoring  & 28.0 & 82.94 & 94.19 & 17.6 & 77.09 & 93.00 \\
 15 & ROVER over S to L  & 31.1 & 83.83 & 66.90 & 18.2 & 77.76 & 65.50 \\
 16 & \quad + N-best rescoring  & 30.0 & 83.38 & 88.57 & 18.7 & 77.32 & 86.50 \\
 \hline
 17 & \textbf{length encoding (tokens)} & 31.5 & 83.91 & 48.57 & 19.6 & 77.45 & 55.50 \\
 18 & \quad + 2-pass length correction  & 30.0 & 83.42 & 68.14 & 19.5 & 77.75 & 75.50 \\
 19 & \quad + N-best rescoring  & 30.9 & 83.66 & 72.36 & 19.3 & 77.47 & 80.50 \\
 20 & \quad \quad + 2-pass length correction  & 29.5 & 83.12 & 88.41 & 19.0 & 76.95 & 92.00 \\
 \hline
 21 & \textbf{length encoding (characters)}  & 30.7 & 83.57 & 63.64 & 20.1 & 78.27 & 73.00 \\
 22 & \quad + 2-pass length correction  & 29.3 & 82.89 & 89.50 & 19.2 & 77.55 & 90.50 \\
 23 & \quad + N-best rescoring  & 30.0 & 83.24 & 88.10 & 19.2 & 77.22 & 95.50 \\
 24 & \quad \quad + 2-pass length correction  & 29.2 & 82.76 & 98.14 & 18.8 & 76.80 & 98.00 \\
 \hline
\end{tabular}
\caption{English$\rightarrow$German translation results for MuST-C tst-COMMON and the IWSLT 2022 Isomtetric SLT blind test. All values in \%. LC = length compliance within 10\% in number of characters. All systems are based on the same Transformer \textit{big} model. Length bins of the 7-bin system are referred to as XXS, XS, S, M, L, XL and XXL from short to long. For explanation of N-best rescoring, ROVER, and 2-pass length correction refer to Section \ref{sec:length_control}.}
\label{table:results}
\end{table*}

\subsection{Forward-translated data}
In addition to back-translated data, we also augmented our training corpus with forward-translated data. For this, we generated translations using our English-to-German system with 7 length bins and a source length token for each of the length classes. Then, we kept only those translations which turned out to be length-compliant with the corresponding source sentence. The resulting synthetic corpus has 213K sentence pairs.

\section{Experimental results}\label{sec:experiments}

Table \ref{table:results} presents results for all length control methods explored in this work. We evaluate on MuST-C tst-COMMON v2\footnote{The official evaluation uses tst-COMMON v1. Differences in metric scores are minor though.} and the blind test set provided by the shared task organizers using the official scoring script\footnote{Blind test set and scoring script are published under \url{https://github.com/amazon-research/isometric-slt}.}. As a measure of MT quality it computes BLEU \citep{papineni-etal-2002-bleu, post2018call} and BERT F1 score \citep{bert-score}. Length compliance (LC) is calculated as the proportion of translations that have a character count which differs by 10\% or less from the number of characters in the source sentence. For this, spaces are not counted and sentences with less than 10 characters are ignored.
References for the blind test set were made available only after development of the systems. Line 0 in Table \ref{table:results} corresponds to a system trained without any of the length control methods from Section \ref{sec:length_control}. All systems use all synthetic data as described in Section \ref{sec:synthetic} if not stated otherwise. 

\subsection{Length token systems}

Rows 1 to 4 of Table \ref{table:results} show results for the 3-bin length token systems. The "length compliant" bin is used for all translations. (When used on the target side it is enforced as the first decoding step.) Overall, we observe no major differences between a source-side and target-side length token in both LC and MT quality scores. Synthetic data and selection of the length bin alone leads to length compliant translations in about 50\% of cases (rows 1 and 3). This shows that the model has to compromise between translation quality and length and that a length token is not a strong enough signal to enforce the corresponding length class in all cases.
N-best rescoring, i.e. selection of a length compliant translation from the beam search output of size 12, can improve LC to 78\% on tst-COMMON but comes at the cost of a loss in translation quality by 0.8\% BLEU and 0.3\% BERTScore absolute.

The 7-bin system shown in rows 5 to 16 offers a greater variety of trade-off points.
We refer to the 7 length bins with size labels from "XXS" to "XXL".
The target-to-source ratio boundaries for equally sized bins in terms of training examples are computed to be 0.90, 0.98, 1.02, 1.06, 1.10, and 1.23. This means the desired 1.0 ratio for isometric MT falls into the "S" bin.

Row 5 shows the scores achieved when not forcing any length token. This configuration leads to the same quality on tst-COMMON as the baseline system, namely 32.0\% BLEU and 84.0\% BERTScore. This indicates that the model is able to predict the right length class corresponding to an unbiased translation. Setting the length token to either "M", "S" or "XS" offers different trade-offs between translation quality and length compliance. Interestingly, the "XS" class has a higher LC than the class "S" which should represent translations with a target-to-source ratio closer to 1. Again, this shows that the effect of length tokens is in conflict with general translation quality, which is optimal when not skipping any information present in the source. A more extreme length class has to be chosen to achieve the desired amount of compression. In all cases N-best rescoring has the same effect as observed for the 3-bin systems, namely a higher LC at the cost of worse translation quality. All length classes not shown in the table lead to either clearly worse LC or quality scores.

The outputs for different length tokens, possibly after N-best rescoring, can be combined with the length ROVER. As mentioned in Section \ref{subsubsec:rover}, we exclude the extreme length classes. We consider two variants: excluding the bins with shortest and longest translations, or excluding the \textit{two} shortest and longest. As expected, both variants lead to more length compliant translations in the combined output. However, they provide different trade-offs: while the first variant (rows 13, 14) can achieve 94\% length compliance on tst-COMMON, translation quality drops to similarly low values as observed for the "XS" length class. The second variant is more conservative and achieves only 89\% length compliance, but preserves higher BLEU and BERT scores.

\subsection{Length encoding systems}
Rows 17 to 24 of Table \ref{table:results} show the results of systems trained with length encoding as described in Section \ref{subsec:length_encoding}. They are also trained using 3 length bins and a "length compliant" token is forced on the target side, we however observe no significant differences to not using the token.

Using the source length as input to the decoder ($L_\mathrm{forced} = L_\mathrm{source}$), the token-level length encoding model (row 17) does not achieve a higher LC value than the length token systems (49\%), while the model with character-level length encoding (row 21) is able to produce compliant translations in 64\% of the cases. Doing a length-corrected second decoding pass is very effective for both systems. This shows that the decoder input $L_\mathrm{forced}$ has a strong impact on the model output, however has to be adjusted to get the desired output length. In Section \ref{subsubsec:length_perturbation} we give explanations for such imperfections. In addition, similar to the case of length tokens, we attribute this to the fact that in training the desired length is always conform with the reference translation, while at inference time the model often has to compress its output to fulfill the length constraints, which might require a more extreme value for the targeted length $L_\mathrm{forced}$.

N-best rescoring can be applied on top to achieve a further large increase in length compliance\footnote{First-best translation length of first pass is used for length correction, N-best rescoring only applied in the second pass.}. This indicates that there is length variety in the N-best list that at least in part can be attributed to the noise added through length perturbation (Section \ref{subsubsec:length_perturbation}). The resulting character-level length encoding system in row 24 achieves the overall best length compliance value of 98.14\%.

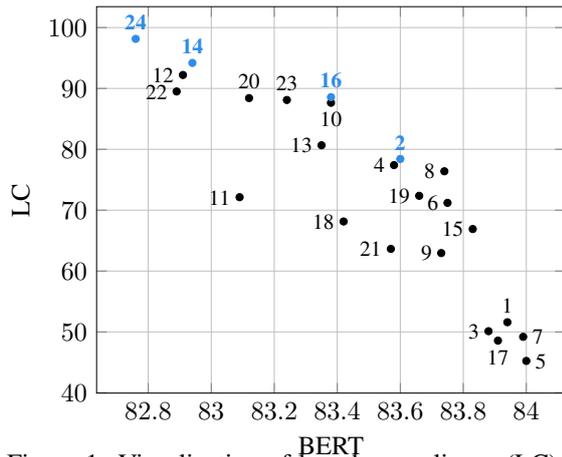
\begin{figure}[h!t]
    \centering
    \definecolor{bleudefrance}{rgb}{0.19, 0.55, 0.91}
\begin{tikzpicture}[scale=0.9]
\hspace{-0.2cm}
\begin{axis}[%
nodes near coords,
ylabel={LC},
xlabel={BERT},
ytick={40,50,60,70,80,90,100},
grid=both]
\addplot[scatter, only marks,%
    scatter/classes={a={black}, b={bleudefrance}},
    point meta=explicit symbolic,
    mark=*,
    mark size=1.5pt,%
    nodes near coords*={\Label},
    visualization depends on={value \thisrow{anchor}\as\myanchor},
    visualization depends on={value \thisrow{color}\as\mycolor},
    visualization depends on={value \thisrow{label}\as\Label},
    every node near coord/.append style={font=\small,anchor=\myanchor,color=\mycolor},%
    ] table[x=x, y=y, meta=class] {
x y label class anchor color
83.94 51.59 1 a {south} {black}
83.60 78.41 \textbf{2} b {south} {bleudefrance}
83.88 50.12 3 a {east} {black}
83.58 77.40 4 a {east} {black}
84.00 45.23 5 a {west} {black}
83.75 71.20 6 a {east} {black}
83.99 49.19 7 a {west} {black}
83.74 76.39 8 a {east} {black}
83.73 62.95 9 a {east} {black}
83.38 87.64 10 a {north} {black}
83.09 72.13 11 a {east} {black}
82.91 92.21 12 a {east} {black}
83.35 80.66 13 a {east} {black}
82.94 94.19 \textbf{14} b {south} {bleudefrance}
83.83 66.90 15 a {east} {black}
83.38 88.57 \textbf{16} b {south} {bleudefrance}
83.91 48.57 17 a {north} {black}
83.42 68.14 18 a {east} {black}
83.66 72.36 19 a {east} {black}
83.12 88.41 20 a {south} {black}
83.57 63.64 21 a {east} {black}
82.89 89.50 22 a {east} {black}
83.24 88.10 23 a {south} {black}
82.76 98.14 \textbf{24} b {south} {bleudefrance}
    };
\end{axis}
\end{tikzpicture}
    \vspace{-1cm}
    \caption{Visualization of length compliance (LC) vs. BERTScore trade-offs on MuST-C tst-COMMON for systems taken from Table \ref{table:results}. Data point labels are the row numbers (\#) from Table \ref{table:results}. Submitted systems are labeled in bold blue.}
    \label{fig:LC_vs_BERT}
\end{figure}

\subsection{System selection}
To select systems for our submission, in
Figure \ref{fig:LC_vs_BERT} we visualize the inherent trade-off between length compliance and translation quality for the systems from Table \ref{table:results}. We look at BERT scores as they were announced to be the main MT quality metric for the evaluation. We chose system 16, the 7-bin length token system using the length ROVER, as our primary submission. As contrastive submissions we include systems 2 (3 length bins using source-side token), 14 (ROVER variation of the primary submission) and 24 (character-level length encoding with second-pass length correction). All submissions use N-best rescoring. As it can be seen, the different length control methods are all able to provide useful trade-off points. While only length encoding can achieve a near perfect length compliance, length token-based methods can offer a good compromise that preserves more of the baseline MT performance.

\subsection{Ablation study}

For a selected subset of the systems we show the contribution of the most important types of synthetic data used in our systems (Section \ref{sec:synthetic}), as well as the effect of length perturbation (Section \ref{subsubsec:length_perturbation}).

\begin{table*}[h!t]
\small
\centering
\begin{tabular}{| c | l | c c c | c c c |}
 \hline
 & & \multicolumn{3}{|c|}{\textbf{tst-COMMON v2}} & \multicolumn{3}{|c|}{\textbf{blind test}} \\
 \# & & \textbf{BLEU} & \textbf{BERT} & \textbf{LC} & \textbf{BLEU} & \textbf{BERT} & \textbf{LC} \\
 \hline
 & \textbf{target-side token, 7 bins} & & & & & & \\
 1 & Row 16, Table \ref{table:results} & 30.0 & 83.38 & 88.57 & 18.7 & 77.32 & 86.50 \\
 2 & \quad + no synonym replacement & 29.6 & 83.41 & 88.41 & 20.0 & 77.58 & 88.50 \\
 3 & \quad \quad + no back-/forward-translation & 29.5 & 83.20 & 87.48 & 19.5 & 77.49 & 87.50 \\
 \hline
 & \textbf{length encoding (tokens)} & & & & & & \\
 4 & Row 19, Table \ref{table:results} & 30.9 & 83.66 & 72.36 & 19.3 & 77.47 & 80.50 \\
 5 & \quad + no length perturbation & 28.6 & 82.32 & 76.12 & 18.3 & 74.51 & 81.00 \\
 \hline
 & \textbf{length encoding (characters)} & & & & & & \\
 6 & Row 21, Table \ref{table:results} & 30.7 & 83.57 & 63.64 & 20.1 & 78.27 & 73.00 \\
 7 & \quad + no length perturbation & 26.6 & 81.66 & 98.26 & 18.4 & 76.07 & 99.00 \\
  & \quad + no synonyms replacement, & & & & & & \\
 8 & \quad \phantom{+} no back-/forward-translation & 30.0 & 83.37 & 61.94 & 19.8 & 77.86 & 75.50 \\
 \hline
\end{tabular}
\caption{Ablation study results. All values in \%.}
\label{table:ablation_results}
\vspace{-0.3cm}
\end{table*}

\subsubsection{Effect of synthetic data}

Comparison of the first two rows of Table \ref{table:ablation_results} shows that taking away synthetic data created using word synonym replacement (Section \ref{subsec:synonyms}) from the 7-bin length token system causes a slight degradation of the BLEU score and no significant change of BERT and length compliance score on tst-COMMON. We consistently observe the same tendencies when taking other configurations of the 7-bin system from Table \ref{table:results} as baseline (not shown here). This indicates that synonym replacement has some positive effect on MT quality as a data augmentation method, but fails to lead to the desired effect of improved length compliance. This could also in part be explained by the fact that in our experiment setting, removing synonym data resulted in the increased relative proportion of length-compliant back- and forward-translated data.

Removing also the back- and forward-translated data from training leads to a consistent drop in all quality metrics on tst-COMMON. In particular, length compliance becomes worse, even in the considered case that uses the length ROVER and N-best rescoring. When training the length-unbiased system of row 5, Table \ref{table:results} without synthetic data LC even drops from 45.27 to 30.70 (not shown in Table \ref{table:ablation_results}). This shows that length-compliant back- and forward-translated data clearly has the desired effect of learning isometric translation and it is still noticeable when combined with other length control methods. Also for the length encoding model (row 8) we observe a similar positive effect of the synthetic data, despite the translation length being predominantly determined by the length value fed into the decoder. 

On the blind test set we observe contradicting results. For this we can provide no better explanation than referring to statistical randomness. In Table \ref{table:results} one can see that ranking of independently trained neural models (e.g. rows 1, 3, 5, 17 and 21) disagrees on the two test sets, which we attribute to the small size of 200 lines of the blind test set. In fact, according to paired bootstrap resampling computed with SacreBLEU \cite{post2018call}, the large difference of 1.3 BLEU between row 1 and 2 of Table \ref{table:ablation_results} is not statistically significant with $p<0.05$, and the 95\% confidence interval of row 1 is 2.8 BLEU.

\subsubsection{Effect of length perturbation}
Without length perturbation the character-level length encoding model is able to produce length compliant translations in almost all cases, as can be seen in Row 7 of Table \ref{table:ablation_results}, without the need for subsequent steps like N-best rescoring or second-pass length correction. This however comes at the cost of a severe drop in translation quality as measured in both BLEU and BERTScore. When comparing to row 24 of Table \ref{table:results} it is apparent that the system trained with length perturbation and using the above-mentioned methods can achieve a similar high level of length compliance while offering a better translation quality by 2.6\% BLEU and 1.1\% BERT F1 score absolute.

A similar drop in translation quality due to lack of length perturbation can be observed for the case of token-level length encoding comparing rows 4 and 5 of Table \ref{table:ablation_results}. The gain in LC from training without noise is outperformed by the combination of N-best rescoring and second-pass length correction applied to the baseline system (row 20, Table \ref{table:results}). Notably, even without noise in training token-level length encoding does not surpass a length compliance value of 80\%. This shows that the number of subwords is not accurate enough as a measure of length when targeting a  precise character count.

\section{Conclusion}\label{sec:conclusions}
In this paper, we described AppTek's neural MT system with length control that we submitted to the IWSLT 2022 Isometric Spoken Translation Evaluation. We showed that by using length-compliant synthetic data, as well as encoding the desired translation length in various ways, we can significantly increase the length compliance score, while at the same time limiting the loss of information as reflected in only slightly lower BERT scores. As one of the best methods for real-time production settings not involving system combination, N-best list rescoring or 2-pass search, the modified positional encoding that counts the desired length in characters achieves the best quality/length compliance trade-off in our experiments. We attribute this to more fine-grained length control capabilities of this system as compared to systems that use source-side or target-side length pseudo-tokens.


\bibliography{anthology,custom}

\begin{thebibliography}{25}
\expandafter\ifx\csname natexlab\endcsname\relax\def\natexlab#1{#1}\fi

\bibitem[{Abadi et~al.(2015)Abadi, Agarwal, Barham, Brevdo, Chen, Citro,
  Corrado, Davis, Dean, Devin, Ghemawat, Goodfellow, Harp, Irving, Isard, Jia,
  Jozefowicz, Kaiser, Kudlur, Levenberg, Man\'{e}, Monga, Moore, Murray, Olah,
  Schuster, Shlens, Steiner, Sutskever, Talwar, Tucker, Vanhoucke, Vasudevan,
  Vi\'{e}gas, Vinyals, Warden, Wattenberg, Wicke, Yu, and
  Zheng}]{tensorflow2015-whitepaper}
Mart\'{i}n Abadi, Ashish Agarwal, Paul Barham, Eugene Brevdo, Zhifeng Chen,
  Craig Citro, Greg~S. Corrado, Andy Davis, Jeffrey Dean, Matthieu Devin,
  Sanjay Ghemawat, Ian Goodfellow, Andrew Harp, Geoffrey Irving, Michael Isard,
  Yangqing Jia, Rafal Jozefowicz, Lukasz Kaiser, Manjunath Kudlur, Josh
  Levenberg, Dandelion Man\'{e}, Rajat Monga, Sherry Moore, Derek Murray, Chris
  Olah, Mike Schuster, Jonathon Shlens, Benoit Steiner, Ilya Sutskever, Kunal
  Talwar, Paul Tucker, Vincent Vanhoucke, Vijay Vasudevan, Fernanda Vi\'{e}gas,
  Oriol Vinyals, Pete Warden, Martin Wattenberg, Martin Wicke, Yuan Yu, and
  Xiaoqiang Zheng. 2015.
\newblock \href {https://www.tensorflow.org/} {{TensorFlow}: Large-scale
  machine learning on heterogeneous systems}.
\newblock Software available from tensorflow.org.

\bibitem[{Akhbardeh et~al.(2021)Akhbardeh, Arkhangorodsky, Biesialska, Bojar,
  Chatterjee, Chaudhary, Costa-jussa, Espa{\~n}a-Bonet, Fan, Federmann,
  Freitag, Graham, Grundkiewicz, Haddow, Harter, Heafield, Homan, Huck,
  Amponsah-Kaakyire, Kasai, Khashabi, Knight, Kocmi, Koehn, Lourie, Monz,
  Morishita, Nagata, Nagesh, Nakazawa, Negri, Pal, Tapo, Turchi, Vydrin, and
  Zampieri}]{akhbardeh-etal-2021-findings}
Farhad Akhbardeh, Arkady Arkhangorodsky, Magdalena Biesialska, Ond{\v{r}}ej
  Bojar, Rajen Chatterjee, Vishrav Chaudhary, Marta~R. Costa-jussa, Cristina
  Espa{\~n}a-Bonet, Angela Fan, Christian Federmann, Markus Freitag, Yvette
  Graham, Roman Grundkiewicz, Barry Haddow, Leonie Harter, Kenneth Heafield,
  Christopher Homan, Matthias Huck, Kwabena Amponsah-Kaakyire, Jungo Kasai,
  Daniel Khashabi, Kevin Knight, Tom Kocmi, Philipp Koehn, Nicholas Lourie,
  Christof Monz, Makoto Morishita, Masaaki Nagata, Ajay Nagesh, Toshiaki
  Nakazawa, Matteo Negri, Santanu Pal, Allahsera~Auguste Tapo, Marco Turchi,
  Valentin Vydrin, and Marcos Zampieri. 2021.
\newblock \href {https://aclanthology.org/2021.wmt-1.1} {Findings of the 2021
  conference on machine translation ({WMT}21)}.
\newblock In \emph{Proceedings of the Sixth Conference on Machine Translation},
  pages 1--88, Online. Association for Computational Linguistics.

\bibitem[{Anastasopoulos et~al.(2022)Anastasopoulos, Bentivogli, Boito, Bojar,
  Cattoni, Currey, Dinu, Duh, Elbayad, Federico, Federmann, Gong, Grundkiewicz,
  Haddow, Hsu, Javorský, Kloudová, Lakew, Ma, Mathur, McNamee, Murray,
  N\u{a}dejde, Nakamura, Negri, Niehues, Niu, Pino, Salesky, Shi, St\"uker,
  Sudoh, Turchi, Virkar, Waibel, Wang, and Watanabe}]{iwslt:2022}
Antonios Anastasopoulos, Luisa Bentivogli, Marcely~Z. Boito, Ondřej Bojar,
  Roldano Cattoni, Anna Currey, Georgiana Dinu, Kevin Duh, Maha Elbayad,
  Marcello Federico, Christian Federmann, Hongyu Gong, Roman Grundkiewicz,
  Barry Haddow, Benjamin Hsu, Dávid Javorský, Věra Kloudová, Surafel~M.
  Lakew, Xutai Ma, Prashant Mathur, Paul McNamee, Kenton Murray, Maria
  N\u{a}dejde, Satoshi Nakamura, Matteo Negri, Jan Niehues, Xing Niu, Juan
  Pino, Elizabeth Salesky, Jiatong Shi, Sebastian St\"uker, Katsuhito Sudoh,
  Marco Turchi, Yogesh Virkar, Alex Waibel, Changhan Wang, and Shinji Watanabe.
  2022.
\newblock {FINDINGS} {OF} {THE} {IWSLT} 2022 {EVALUATION} {CAMPAIGN}.
\newblock In \emph{Proceedings of the 19th International Conference on Spoken
  Language Translation (IWSLT 2022)}, Dublin, Ireland. Association for
  Computational Linguistics.

\bibitem[{Buet and Yvon(2021)}]{buet2021toward}
Fran{\c{c}}ois Buet and Fran{\c{c}}ois Yvon. 2021.
\newblock \href
  {https://www.isca-speech.org/archive/pdfs/interspeech_2021/buet21_interspeech.pdf}
  {Toward genre adapted closed captioning}.
\newblock In \emph{Interspeech 2021}, pages 4403--4407. ISCA.

\bibitem[{Di~Gangi et~al.(2019)Di~Gangi, Cattoni, Bentivogli, Negri, and
  Turchi}]{di2019must}
Mattia~A Di~Gangi, Roldano Cattoni, Luisa Bentivogli, Matteo Negri, and Marco
  Turchi. 2019.
\newblock \href {https://aclanthology.org/N19-1202/} {Must-c: a multilingual
  speech translation corpus}.
\newblock In \emph{Proceedings of the 2019 Conference of the North American
  Chapter of the Association for Computational Linguistics: Human Language
  Technologies, Volume 1 (Long and Short Papers)}, pages 2012--2017.

\bibitem[{Doetsch et~al.(2017)Doetsch, Zeyer, Voigtlaender, Kulikov,
  Schl{\"u}ter, and Ney}]{doetsch2017returnn}
Patrick Doetsch, Albert Zeyer, Paul Voigtlaender, Ilia Kulikov, Ralf
  Schl{\"u}ter, and Hermann Ney. 2017.
\newblock \href
  {http://web-info8.informatik.rwth-aachen.de/media/papers/0005345.pdf}
  {Returnn: The rwth extensible training framework for universal recurrent
  neural networks}.
\newblock In \emph{2017 IEEE International Conference on Acoustics, Speech and
  Signal Processing (ICASSP)}, pages 5345--5349. IEEE.

\bibitem[{Dyer et~al.(2013)Dyer, Chahuneau, and Smith}]{dyer-etal-2013-simple}
Chris Dyer, Victor Chahuneau, and Noah~A. Smith. 2013.
\newblock \href {https://aclanthology.org/N13-1073} {A simple, fast, and
  effective reparameterization of {IBM} model 2}.
\newblock In \emph{Proceedings of the 2013 Conference of the North {A}merican
  Chapter of the Association for Computational Linguistics: Human Language
  Technologies}, pages 644--648, Atlanta, Georgia. Association for
  Computational Linguistics.

\bibitem[{Fiscus(1997)}]{ROVER}
J.G. Fiscus. 1997.
\newblock \href {https://doi.org/10.1109/ASRU.1997.659110} {A post-processing
  system to yield reduced word error rates: Recognizer output voting error
  reduction (rover)}.
\newblock In \emph{1997 IEEE Workshop on Automatic Speech Recognition and
  Understanding Proceedings}, pages 347--354.

\bibitem[{Garc{\'\i}a-Mart{\'\i}nez et~al.(2016)Garc{\'\i}a-Mart{\'\i}nez,
  Barrault, and Bougares}]{garcia-martinez-etal-2016-factored}
Mercedes Garc{\'\i}a-Mart{\'\i}nez, Lo{\"\i}c Barrault, and Fethi Bougares.
  2016.
\newblock \href {https://aclanthology.org/2016.iwslt-1.3} {Factored neural
  machine translation architectures}.
\newblock In \emph{Proceedings of the 13th International Conference on Spoken
  Language Translation}, Seattle, Washington D.C. International Workshop on
  Spoken Language Translation.

\bibitem[{Kudo and Richardson(2018)}]{kudo-richardson-2018-sentencepiece}
Taku Kudo and John Richardson. 2018.
\newblock \href {https://doi.org/10.18653/v1/D18-2012} {{S}entence{P}iece: A
  simple and language independent subword tokenizer and detokenizer for neural
  text processing}.
\newblock In \emph{Proceedings of the 2018 Conference on Empirical Methods in
  Natural Language Processing: System Demonstrations}, pages 66--71, Brussels,
  Belgium. Association for Computational Linguistics.

\bibitem[{Lakew et~al.(2021)Lakew, Federico, Wang, Hoang, Virkar,
  Barra-Chicote, and Enyedi}]{lakew2021machine}
Surafel~M Lakew, Marcello Federico, Yue Wang, Cuong Hoang, Yogesh Virkar,
  Roberto Barra-Chicote, and Robert Enyedi. 2021.
\newblock \href {https://arxiv.org/pdf/2110.03847.pdf} {Machine translation
  verbosity control for automatic dubbing}.
\newblock In \emph{ICASSP 2021-2021 IEEE International Conference on Acoustics,
  Speech and Signal Processing (ICASSP)}, pages 7538--7542. IEEE.

\bibitem[{Lakew et~al.(2019)Lakew, Di~Gangi, and
  Federico}]{lakew-etal-2019-controlling}
Surafel~Melaku Lakew, Mattia Di~Gangi, and Marcello Federico. 2019.
\newblock \href {https://aclanthology.org/2019.iwslt-1.31} {Controlling the
  output length of neural machine translation}.
\newblock In \emph{Proceedings of the 16th International Conference on Spoken
  Language Translation}, Hong Kong. Association for Computational Linguistics.

\bibitem[{Liu et~al.(2020)Liu, Niehues, and Spanakis}]{liu-etal-2020-adapting}
Danni Liu, Jan Niehues, and Gerasimos Spanakis. 2020.
\newblock \href {https://doi.org/10.18653/v1/2020.iwslt-1.30} {Adapting
  end-to-end speech recognition for readable subtitles}.
\newblock In \emph{Proceedings of the 17th International Conference on Spoken
  Language Translation}, pages 247--256, Online. Association for Computational
  Linguistics.

\bibitem[{Niehues(2020)}]{niehues-2020-machine}
Jan Niehues. 2020.
\newblock \href {https://aclanthology.org/2020.amta-research.3} {Machine
  translation with unsupervised length-constraints}.
\newblock In \emph{Proceedings of the 14th Conference of the Association for
  Machine Translation in the Americas (Volume 1: Research Track)}, pages
  21--35, Virtual. Association for Machine Translation in the Americas.

\bibitem[{Oka et~al.(2020)Oka, Chousa, Sudoh, and
  Nakamura}]{oka-etal-2020-incorporating}
Yui Oka, Katsuki Chousa, Katsuhito Sudoh, and Satoshi Nakamura. 2020.
\newblock \href {https://doi.org/10.18653/v1/2020.coling-main.319}
  {Incorporating noisy length constraints into transformer with length-aware
  positional encodings}.
\newblock In \emph{Proceedings of the 28th International Conference on
  Computational Linguistics}, pages 3580--3585, Barcelona, Spain (Online).
  International Committee on Computational Linguistics.

\bibitem[{Papineni et~al.(2002)Papineni, Roukos, Ward, and
  Zhu}]{papineni-etal-2002-bleu}
Kishore Papineni, Salim Roukos, Todd Ward, and Wei-Jing Zhu. 2002.
\newblock \href {https://doi.org/10.3115/1073083.1073135} {{B}leu: a method for
  automatic evaluation of machine translation}.
\newblock In \emph{Proceedings of the 40th Annual Meeting of the Association
  for Computational Linguistics}, pages 311--318, Philadelphia, Pennsylvania,
  USA. Association for Computational Linguistics.

\bibitem[{Peter et~al.(2018)Peter, Beck, and Ney}]{peter-etal-2018-sisyphus}
Jan-Thorsten Peter, Eugen Beck, and Hermann Ney. 2018.
\newblock \href {https://doi.org/10.18653/v1/D18-2015} {Sisyphus, a workflow
  manager designed for machine translation and automatic speech recognition}.
\newblock In \emph{Proceedings of the 2018 Conference on Empirical Methods in
  Natural Language Processing: System Demonstrations}, pages 84--89, Brussels,
  Belgium. Association for Computational Linguistics.

\bibitem[{Post(2018)}]{post2018call}
Matt Post. 2018.
\newblock \href {https://www.statmt.org/wmt18/pdf/WMT019.pdf} {A call for
  clarity in reporting bleu scores}.
\newblock In \emph{Proceedings of the Third Conference on Machine Translation:
  Research Papers}, pages 186--191.

\bibitem[{Saboo and Baumann(2019)}]{saboo-baumann-2019-integration}
Ashutosh Saboo and Timo Baumann. 2019.
\newblock \href {https://doi.org/10.18653/v1/W19-5210} {Integration of dubbing
  constraints into machine translation}.
\newblock In \emph{Proceedings of the Fourth Conference on Machine Translation
  (Volume 1: Research Papers)}, pages 94--101, Florence, Italy. Association for
  Computational Linguistics.

\bibitem[{Sennrich et~al.(2016)Sennrich, Haddow, and
  Birch}]{sennrich-etal-2016-neural}
Rico Sennrich, Barry Haddow, and Alexandra Birch. 2016.
\newblock \href {https://doi.org/10.18653/v1/P16-1162} {Neural machine
  translation of rare words with subword units}.
\newblock In \emph{Proceedings of the 54th Annual Meeting of the Association
  for Computational Linguistics (Volume 1: Long Papers)}, pages 1715--1725,
  Berlin, Germany. Association for Computational Linguistics.

\bibitem[{Takase and Okazaki(2019)}]{takase-okazaki-2019-positional}
Sho Takase and Naoaki Okazaki. 2019.
\newblock \href {https://doi.org/10.18653/v1/N19-1401} {Positional encoding to
  control output sequence length}.
\newblock In \emph{Proceedings of the 2019 Conference of the North {A}merican
  Chapter of the Association for Computational Linguistics: Human Language
  Technologies, Volume 1 (Long and Short Papers)}, pages 3999--4004,
  Minneapolis, Minnesota. Association for Computational Linguistics.

\bibitem[{Vaswani et~al.(2017)Vaswani, Shazeer, Parmar, Uszkoreit, Jones,
  Gomez, Kaiser, and Polosukhin}]{vaswani2017attention}
Ashish Vaswani, Noam Shazeer, Niki Parmar, Jakob Uszkoreit, Llion Jones,
  Aidan~N Gomez, {\L}ukasz Kaiser, and Illia Polosukhin. 2017.
\newblock \href
  {https://proceedings.neurips.cc/paper/2017/file/3f5ee243547dee91fbd053c1c4a845aa-Paper.pdf}
  {Attention is all you need}.
\newblock \emph{Advances in neural information processing systems}, 30.

\bibitem[{Wilken and Matusov(2019)}]{wilken2019novel}
Patrick Wilken and Evgeny Matusov. 2019.
\newblock \href {https://arxiv.org/pdf/1910.03912.pdf} {Novel applications of
  factored neural machine translation}.
\newblock \emph{arXiv preprint arXiv:1910.03912}.

\bibitem[{Zeyer et~al.(2018)Zeyer, Alkhouli, and Ney}]{zeyer-etal-2018-returnn}
Albert Zeyer, Tamer Alkhouli, and Hermann Ney. 2018.
\newblock \href {https://doi.org/10.18653/v1/P18-4022} {{RETURNN} as a generic
  flexible neural toolkit with application to translation and speech
  recognition}.
\newblock In \emph{Proceedings of {ACL} 2018, System Demonstrations}, pages
  128--133, Melbourne, Australia. Association for Computational Linguistics.

\bibitem[{Zhang et~al.(2020)Zhang, Kishore, Wu, Weinberger, and
  Artzi}]{bert-score}
Tianyi Zhang, Varsha Kishore, Felix Wu, Kilian~Q. Weinberger, and Yoav Artzi.
  2020.
\newblock \href {https://openreview.net/forum?id=SkeHuCVFDr} {Bertscore:
  Evaluating text generation with bert}.
\newblock In \emph{International Conference on Learning Representations}.

\end{thebibliography}
\bibliographystyle{acl_natbib}


\end{document}